# Efficient Malicious UAV Detection Using Autoencoder-TSMamba Integration

Azim Akhtarshenas[1][0000−0002−5511−1010], Ramin Toosi[2], David López-Pérez[3], Tohid Alizadeh[4], and Alireza Hosseini[5]

[1] Universitat Politècnica de València, Valencia, Spain, aakhtar@doctor.upv.es
[2] University of Tehran, Tehran, Iran, r.toosi@ut.ac.ir
[3] Universitat Politècnica de València, Valencia, Spain, D.lopez@iteam.upv.es
[4] Nazarbayev University, Astana, Kazakhstan, tohid.alizadeh@nu.edu.kz
[5] University of Tehran, Tehran, Iran, arhosseini77@ut.ac.ir

**Abstract.** Malicious Unmanned Aerial Vehicles (UAVs) present a significant threat to next-generation networks (NGNs), posing risks such as unauthorized surveillance, data theft, and the delivery of hazardous materials. This paper proposes an integrated (AE)-classifier system to detect malicious UAVs. The proposed AE, based on a 4-layer Tri-orientated Spatial Mamba (TSMamba) architecture, effectively captures complex spatial relationships crucial for identifying malicious UAV activities. The first phase involves generating residual values through the AE, which are subsequently processed by a ResNet-based classifier. This classifier leverages the residual values to achieve lower complexity and higher accuracy. Our experiments demonstrate significant improvements in both binary and multi-class classification scenarios, achieving up to 99.8 % recall compared to 96.7 % in the benchmark. Additionally, our method reduces computational complexity, making it more suitable for large-scale deployment. These results highlight the robustness and scalability of our approach, offering an effective solution for malicious UAV detection in NGN environments.

## 1  Introduction

The rapid development of next-generation networks (NGNs), including advancements such as 5G and beyond, combined with the integration of unmanned aerial vehicles (UAVs), has unlocked remarkable opportunities for connectivity and innovation. UAVs, commonly known as drones, have vast potential across various industries and applications, including agriculture, surveillance, logistics, and entertainment. However, their widespread availability and versatility also raise significant concerns regarding potential misuse for malicious purposes, which could pose serious threats to airspace and society [1]. To address the threat posed by malicious UAVs, the authors in [2] survey the key features of such UAVs, potential threats, and current countermeasures, including detection, tracking, and classification methods. They also discuss the limitations of existing approaches in the literature, and suggest future research directions to improve the management of UAV threats. One conclusion drawn from this and similar studies is that, to prevent harm, it is imperative to design an autonomous platform capable of effectively detecting malevolent UAVs. Among the various methods for detecting malicious UAVs, two prominent approaches are communication-based methods [1, 3–5] and computer vision-based methods [1, 6, 7], the latter largely relying on artificial intelligence (AI) tools. In the following, we provide a literature review on the topic and highlight the motivation and contribution of our research.

### 1.1 Related Works

**Communication-based methods** The authors in [4] propose a received signal strength (RSS)-based method to detect malicious UAVs in a Rician fading channel. This method leverages RSS variations to identify UAVs in different channel conditions. It uses threshold-based detection, comparing the RSS of the UAV to optimized non-line-of-sight (NLoS) and line-of-sight (LoS) thresholds. In [5], the proposed system detects and tracks drones using ID tags in radio signals, decoding telemetry packets to extract and validate data such as position, altitude, and speed to intercept harmful drones. Using a similar principle, the authors in [8] employ encrypted communication to identify detected UAVs. If identified as friendly, the system decrypts the communication; if not, it reports the UAV to the base station. In [9], MaDe is introduced —a method where each UAV generates an authentication variable based on transmitted packets. These variables are verified by a central device, which uses a generalized likelihood ratio test to detect malicious UAVs. In [10], the authors develop a malicious UAV detection method using a support vector machine (SVM) combined with a shuffled frog leap algorithm. Sensor nodes collect data and transmit it, along with a feedback packet, to the UAV. The SVM-SFL approach classifies the data and optimizes the SVM's performance, improving detection accuracy. The defense system in [11] uses a swarm of self-organizing UAVs to intercept rogue drones. The system employs encrypted communication to identify UAVs and reports threats to the base station, ensuring robust operation. The modular design of this system enhances its scalability and adaptability in dynamic environments. In [12], the authors evaluate swarm formation methods for tracking malicious UAVs. The framework assesses the impact of swarm size and UAV evasiveness on tracking effectiveness. This allows for optimizing swarm strategies to improve the detection and tracking of malicious UAVs. Lastly, [13] discusses the need to integrate UAV flight zones in smart cities and proposes a defense system using self-organizing drone swarms to intercept malicious UAVs. The system uses an auto-balanced clustering mechanism for efficient defense, ensuring resilience against communication losses. This approach promotes coordination among UAVs, improving both detection and mitigation capabilities.

**Computer vision-based methods** In [6], the authors introduced a structure based on a vision transformer (ViT) to discriminate malicious UAVs. Their ap-proach involves segmenting drone images into fixed-size patches. Position and linear embeddings are then incorporated, forming a sequence of vectors processed through a standard ViT encoder. For classification, an additional learnable classification token linked to the sequence is employed. A different solution is suggested in [14], which involves a smartphone application where users can report malicious UAVs, including a photo for identification purposes. The app automatically determines the UAV's manufacturer and specific model using trained image classification models. Four convolutional neural network (CNN) models — AlexNet, VGG-16, ResNet-18, and MobileNet-v2— were trained using a dataset of images from three popular UAVs, captured at various elevations, distances, and camera zoom levels. The study in [15] explores autonomous UAV detection for counter-unmanned aerial systems (CUAS). It employs deep learning models trained on image and acoustic features, combining visual and audio-based methods for enhanced performance. Data collection involved two drones flying simultaneously at a fixed distance, facilitating optimal performance assessment. CNN and YOLOv5 were used for acoustic and visual data analysis, respectively, contributing to a comprehensive drone detection system. In [7], a novel framework combining handcrafted and deep features detects drones using sound and image data, with SVM classifiers applied. Various CNNs and handcrafted descriptors were compared to detect and localize malicious UAVs. According to the authors, combining mel-frequency cepstral coefficients (MFCC) and AlexNet features makes the model robust enough for national security applications. **As the most recent methods**, Autoencoders (AEs) are powerful neural network models used for data reconstruction, encryption, and feature learning. The authors in [16] developed an ensemble approach for intrusion detection by combining AEs and isolation forests. This approach merges the anomaly detection strengths of isolation forests with the feature-learning capabilities of AEs, which excel at learning representations of complex data. In [17], AEs are used to address the challenge of underwater image classification in an open-set scenario. The authors leverage an AE to effectively distinguish between seen and unseen species, improving the model's ability to handle novel data. The authors in [18–20] identified that existing AE frameworks struggle with content-based reasoning and proposed several improvements. To better handle discrete modalities, they transform the state-space model parameters into functions of the input, develop a hardware-optimized parallel algorithm for recurrent processing, and create a streamlined architecture called Mamba that eliminates attention and multi-layer perceptron (MLP) blocks. Mamba scales linearly with sequence length, and excels at handling sequences up to millions of entries. It achieves state-of-the-art performance across various modalities, including language, audio, and genomics. Detailed explanations of Mamba will be provided in the next section, as our proposal builds on it.

## 1.2  Motivation and Contribution

Despite frequent news reports highlighting incidents involving malicious UAVs, current methodologies and datasets to address this issue remain limited, often characterized by complexity or lack of accuracy. This gap poses a significant risk as the potentially harmful applications of UAVs continue to rise. In response to this critical need, and motivated by the work in [6, 7, 18, 19], our paper proposes the design of a CNN-Mamba structure specifically tailored to identify malicious UAVs with high accuracy and reduced complexity Our approach overcomes the limitations of existing methods with two key innovations, as follows:

- **Reducing Model Complexity:** Unlike existing studies that utilize highly complex transformers, our method employs the state-space model introduced by the Mamba structure, which offers lower complexity while achieving higher accuracy.
- **Improved Detection Accuracy:** We focus on residual modeling to enhance accuracy, achieving 100 % recall precision on selected datasets, which is critical for real-world scenarios.

## 2    Proposed Network Structure

Our model process consists of two primary phases: reconstruction and classification. In the reconstruction phase, we calculate the residuals, which are the differences between the initial data and the reconstructed data. During the classification phase, these residual values are fed into the classifier to make final decisions regarding the UAVs Fig. 1 illustrates our pipeline. We present a detailed overview of our system model, covering the datasets and the training and testing processes for both the autoencoder and the classifier.

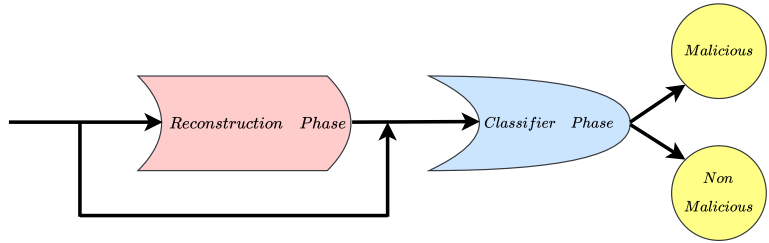

Fig. 1: The block diagram of the proposed method [19].

### 2.1    Dataset

We use the dataset from [6], which comprises 776 images categorized into five classes: kites, drones, malicious drones, birds, balloons, and airplanes. 70 % of the dataset is used for training, while the remaining 30 % is used for validation.

To simulate real-world conditions, the images were captured at varying altitudes (both high and low) and under different weather conditions (favorable and adverse). The dataset also includes variations in brightness, scale, and resolution, further enhancing its value for model training. The diverse conditions ensure a robust training process. Due to limitations in our dataset, we were unable to perform hyperparameter tuning and instead relied on only two datasets: one for training and one for testing.

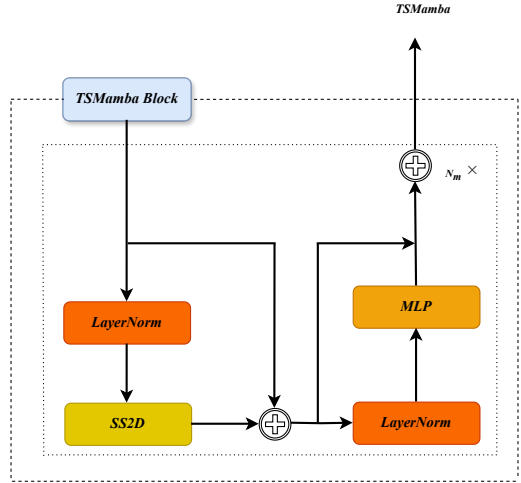

Fig. 2: The structure of TSMamba [19].

## 2.2 Autoencoder

Our model consists of two main phases: reconstruction and classification. In the reconstruction phase, we use an AE to rebuild the input image. The AE takes a 2D input image, and produces a reconstructed version, calculating the difference, or residual, between the original and reconstructed images. This residual helps us identify how well the AE has learned the data. We focus on the classification of malicious drones versus non-malicious drones in this task. Malicious drone datasets tend to be more uniform and well-defined, whereas non-malicious drones encompass a broader variety of types. Due to this, we trained the AE exclusively on malicious drone data, making it highly effective at reconstructing this class. When the AE encounters data from non-malicious drones, its performance drops, which results in high residual values. These residuals serve as inputs for the next step—classification. To further enhance the AE's capabilities, we use TSMamba blocks based on Mamba architecture, a sequence modeling structure that improves upon traditional methods like transformers. Mamba integrates Structured

State Space (S4) models, offering lower complexity and better accuracy by efficiently handling time-varying data. The architecture includes a mechanism that dynamically adjusts the state space model's parameters based on input, focusing on the most relevant data and improving computational efficiency. This makes it ideal for large, sequence-based datasets, like the one we are using for drone detection (see Fig. 2). Each TSMamba block consists of three parts: LayerNorm, SS2D, and MLP. LayerNorm normalizes the input data, which helps stabilize and speed up the training process. SS2D is a 2D version of the Mamba state space, enabling it to process images. In this paper, we propose a 4-layer TS-Mamba model to replace traditional transformer architectures, as shown in Fig. 3. Finally, in the decoder section, we use convolutional layers and upsampling layers to reconstruct the original image from the encoded data. Feature maps from encoding are incorporated during decoding to enhance reconstruction by leveraging local properties learned earlier.

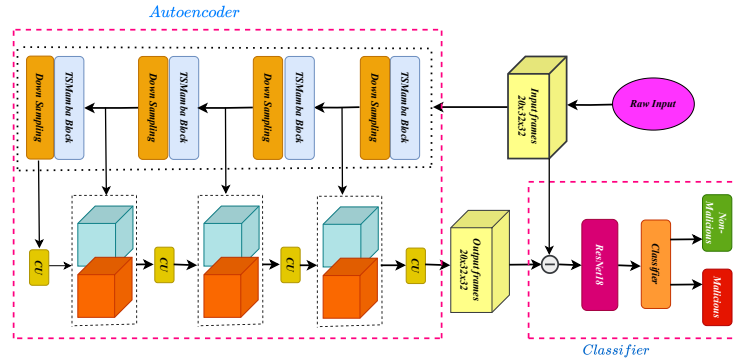

Fig. 3: The detailed structure of the proposed method.

## 2.3   Classifier

After calculating the residual value by subtracting the AE's prediction from the original input in the reconstruction phase, we transition to the second phase of the proposed model. In this phase, using transfer learning, the residual value is fed into a standard ResNet18 model, pre-trained on ImageNet. Since our application requires classifying the images into two categories, we modify the ResNet18 model, which typically has 1,000 output neurons, to have only two output neurons. This modification allows the model to classify the input as either malicious or non-malicious (see Fig. 3).

## 2.4   Key Performance Indicators (KPIs)

To evaluate model performance, we use precision, recall, F1-score, and accuracy.

– **Precision** measures the proportion of correctly predicted positive instances out of all predicted positives:

$$\text{Precision} = \frac{TP}{TP + FP};$$

– **Recall** evaluates the proportion of correctly predicted positives out of all actual positives:

$$\text{Recall} = \frac{TP}{TP + FN};$$

– The **F1-score** is the harmonic mean of precision and recall, balancing their trade-offs:

$$\text{F1} = 2 \cdot \frac{\text{Precision} \cdot \text{Recall}}{\text{Precision} + \text{Recall}};$$

– **Accuracy** is defined as follows:

$$\text{Accuracy} = \frac{TP + TN}{TP + TN + FP + FN};$$

Note that $TP$ refers to the number of true positives (correctly identified positive instances), $FP$ refers to false positives (incorrectly predicted positives), and $FN$ refers to false negatives (positives that were missed by the model).

## 3   Proposed Model Complexity

The computational complexity of Transformers primarily arises from their attention mechanism, which scales quadratically with the input sequence length $n$, resulting in $O(n^2 \cdot d + n \cdot d^2)$, where $d$ is the model dimension. This quadratic dependency makes Transformers Processing-heavy for long sequences. In contrast, **Mamba+CNN (ResNet)** combines Mamba's State Space Model (SSM)-based linear scaling $O(n)$ with the ResNet's convolutional complexity $O(q \cdot k^2 \cdot d \cdot h \cdot w)$, where $q$ is the number of filters, $k$ is the kernel size, $d$ is the depth of the input (number of channels), and $h$ and $w$ are the height and width of the input image. The combined complexity becomes $O(n + q \cdot k^2 \cdot d \cdot h \cdot w)$, which scales linearly with the sequence length $n$, and is significantly more efficient for long sequences compared to the Transformers' quadratic growth. Thus, **Mamba+CNN (ResNet)** is particularly suitable for tasks that require both long-range dependencies and efficient computation, achieving a balance between the capabilities of Mamba for sequence modeling and CNN (ResNet) for spatial feature extraction.

## 4   Experimental and Simulation Results

In this section, we present our experimental and simulation results. The experiment is implemented using the PyTorch deep learning framework, and the computations are accelerated using an NVIDIA RTX 2080 Ti GPU. We also utilized the Adam optimizer with a learning rate of $1.5 \times 10^{-5}$ and conducted the training over 110 epochs. To demonstrate the superiority of our proposed model, we compare its complexity and accuracy metrics with those of [6], which is based on vision transformers. Regarding complexity, their model consists of 82 million parameters, while our proposed model uses only 11.4 million parameters. In the initial part of our experiments, we present the reconstruction loss for the reconstruction phase and the classifier loss and accuracy for the classification phase. The reconstruction loss of the AE during the training and validation

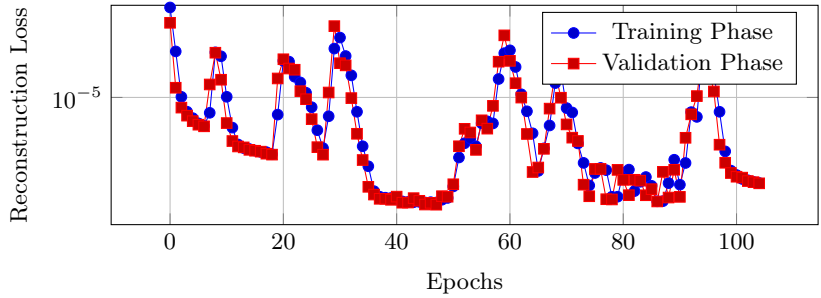

Fig. 4: Logarithmic reconstruction loss in training and validation phases.

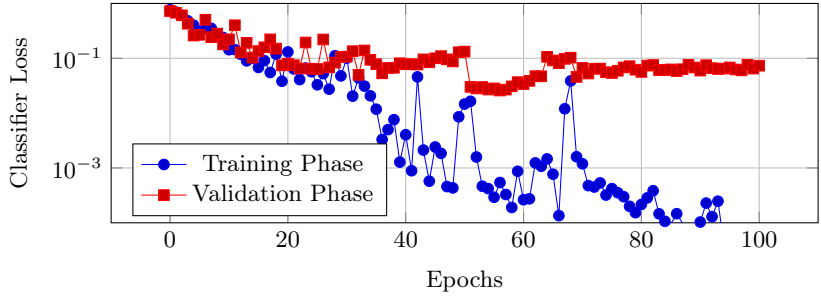

Fig. 5: Logarithmic classifier loss in training and validation phases.

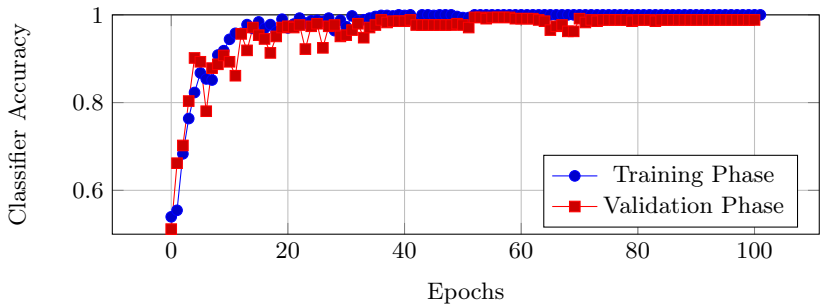

Fig. 6: Classifier accuracy in training and validation phases.

phases for the anomaly class is shown in Fig. 4. A comparison of the training and validation curves reveals a close alignment, indicating that the model is generalizing effectively. This observation underscores the AE's capability to learn meaningful representations, particularly for the anomaly class, and to accurately reconstruct the input data. Fig. 5 and 6 provide a comprehensive view of the classifier's performance during the second phase of training. Specifically, Fig. 5 shows the classification loss over the training epochs, offering insights into how effectively the model reduces errors as it learns to classify the data. After approximately 30 epochs, the loss starts to stabilize, indicating a divergence between the training and validation curves. This suggests that the model has reached a point of diminishing returns, where it has effectively learned from the training data, and the further learning in subsequent epochs do not significantly improve its predictive accuracy. In parallel, Fig. 6 showcases the classification accuracy, highlighting the proportion of correctly classified instances out of the total instances. This figure serves as a direct measure of the model's efficacy in accurately predicting the class labels. By observing the trends in accuracy over the training period, we can envision the model's progression towards better performance after 20 epochs. The alignment of the testing and training curves indicates that the proposed model is generalizes well. In the second part of our

Table 1: Results for methods in [6] and Proposed Method for the five-class scenario, P: Precision (%), R: Recall (%), and F1: F1-score (%)

| Class | Aeroplane | | | Bird | | | Drone | | | Helicopter | | | Malicious Drones | | | Accuracy |
|---|---|---|---|---|---|---|---|---|---|---|---|---|---|---|---|---|
| Method | P | R | F1 | P | R | F1 | P | R | F1 | P | R | F1 | P | R | F1 | % |
| Proposed in [6] | 100 | 100 | 100 | 100 | 100 | 100 | 96 | 100 | 97 | 100 | 100 | 100 | 96 | 100 | 97 | 98 |
| **Our Proposed** | 100 | 100 | 100 | 100 | 100 | 100 | 98 | 100 | 99 | 100 | 100 | 100 | 100 | 98 | 99 | 99.5 |

experiments, we evaluate accuracy, recall, and precision for both multi-class (five-class) and binary classification scenarios. In order to ensure a fair comparison and benchmarking with the reference paper [6], we conducted both two-class (malicious and unmalicious) and five-class classification experiments. Table 1 highlights the significant performance improvements of our proposed method over the approach presented in [6]. Notably, our method achieves near-perfect recall of 99.8 %, compared to 96.7 % achieved by the benchmark across various classes. This demonstrates that our approach is more effective at identifying all relevant instances. Additionally, our method outperforms the benchmark in terms of precision, which reflects the accuracy of the identified instances. This suggests that our model produces fewer false positives, resulting in a more reliable identification process. Table 2 shows the confusion matrix, providing a detailed overview of the model's classification performance in the five-class classification scenario. The diagonal entries, correct classifications, are significantly higher than the off-diagonal entries, which correspond to misclassifications. However, an exception is observed in the drone class, where some misclassification occurs. This suggests that, while the model performs well in distinguishing between most categories,

Table 2: The confusion matrix of the proposed method in five-class scenario.

| | Aeroplane | Birds | Drones | Helicopters | Malicious Drones |
|---|---|---|---|---|---|
| Aeroplane | 33 | 0 | 0 | 0 | 0 |
| Birds | 0 | 31 | 0 | 0 | 0 |
| Drones | 0 | 0 | 59 | 0 | 0 |
| Helicopters | 0 | 0 | 0 | 41 | 0 |
| Malicious Drones | 0 | 0 | 1 | 0 | 67 |

**Accuracy:** 99.5%

it faces challenges in accurately classifying instances of the drone class, likely due to their higher similarity with malicious drones. Table 3 presents the KPIs of our proposed method compared to the approach in [6] for binary classification tasks (distinguishing between malicious drones and non-malicious classes such as drones, birds, helicopters, and airplanes). As shown, our proposed method outperforms the benchmark, achieving improvements of 3 %, 2 %, and 1.5 % in recall, precision, and accuracy, respectively. Furthermore, our method achieves near-perfect recall of 100 %, compared to 96.7 % in the benchmark across binary classes. This superior performance is attributed to our method's unique combination of AE-based residual error analysis with ResNet18 classification.

Table 3: Accuracy, Recall, and Precision comparison for the two-class scenario.

| Comparison Metric | Method in [6] | Our Proposed Method |
|---|---|---|
| **Recall** | 96.7% | **100**% |
| **Precision** | 96.7% | **98.78**% |
| **Accuracy** | 98.2% | **99.42**% |

The autoencoder highlights subtle, malicious-specific features by focusing on reconstruction errors, which serve as discriminative signals for classification. This two-phase approach leverages the strengths of anomaly detection and advanced feature extraction, providing higher sensitivity to malicious patterns.

**Acknowledgement** This research is supported by the Generalitat Valenciana through the CIDEGENT PlaGenT, Grant CIDEXG/2022/17, Project iTENTE.

**Disclosure of Interests.** The authors have no competing interests to declare that are relevant to the content of this article.

## 5 Conclusion

To address the malicious UAV challenges, our paper proposes an integrated AE-classifier system for detecting malicious UAVs. The AE, based on a 4-layer TSMamba architecture, effectively captures complex spatial relationships critical for identifying malicious UAV activities. In the first phase, the AE generates a residual value, which is then fed into the classifier in the second phase. This classifier leverages the residual values to achieve lower complexity and higher accuracy. Our simulation results, which encompass both binary and multi-class (five-class) classification scenarios, demonstrate improvements over the state-of-the-art methods based on Transformers. Specifically, our approach achieves a recall of 99.8 %, compared to 96.7 % in the benchmark, in multi-class scenarios. Additionally, our model reduces computational complexity, making it more efficient for large-scale deployment.